\documentclass[letterpaper, 10 pt, conference]{ieeeconf}

\usepackage{times}
\usepackage{epsfig}
\usepackage{graphicx}
\usepackage{amsmath}
\usepackage{amssymb}

\renewcommand{\baselinestretch}{.98}

\usepackage[noadjust]{cite}

\usepackage[utf8]{inputenc} 
\usepackage[T1]{fontenc}    
\usepackage[pagebackref=true,breaklinks=true,letterpaper=true,colorlinks,bookmarks=false,linkcolor=red,urlcolor=blue]{hyperref}
\usepackage{url}            
\usepackage{booktabs}       
\usepackage{amsfonts}       
\usepackage{nicefrac}       
\usepackage{microtype}      
\usepackage{makecell}
\usepackage{xspace}
\usepackage{paralist}
\usepackage{comment}
\usepackage{mathtools}
\usepackage{xcolor}
\usepackage{ragged2e}
\usepackage{subcaption}
\usepackage[font=small]{caption}
\usepackage{commath}
\usepackage[capitalise,noabbrev,nameinlink]{cleveref}
\usepackage{gensymb}
\usepackage{booktabs}
\usepackage{xparse}
\usepackage{mathtools}
\usepackage{etoolbox}
\usepackage{setspace}
\usepackage{float}
\usepackage{algorithm}
\usepackage[noend]{algpseudocode}
\usepackage{balance}
\usepackage{soul}
\usepackage{bm}

\graphicspath{{imgs/}}

\renewcommand{\norm}[1]{\left\lVert#1\right\rVert}

\DeclarePairedDelimiterX{\SquareBrackets}[1]{[}{]}{#1}
\DeclarePairedDelimiterX{\RoundBrackets}[1]{(}{)}{#1}
\DeclarePairedDelimiterX{\DivergenceBrackets}[2]{[}{]}{#1\;\delimsize\|\;#2}

\NewDocumentCommand{\pr}{ O{p} r() }{
  \def\prArg{#2}\patchcmd{\prArg}{|}{\mid}{}{}#1\RoundBrackets{\prArg}}
\NewDocumentCommand{\p}{ r() }{\pr[p](#1)}
\NewDocumentCommand{\q}{ r() }{\pr[q](#1)}
\NewDocumentCommand{\Normal}{ r() }{\pr[\operatorname{Normal}](#1)}
\NewDocumentCommand{\Cat}{ r() }{\pr[\operatorname{Cat}](#1)}
\NewDocumentCommand{\Bin}{ r() }{\pr[\operatorname{Bin}](#1)}
\NewDocumentCommand{\Beta}{ r() }{\pr[\operatorname{Beta}](#1)}
\NewDocumentCommand{\Bernoulli}{ r() }{\pr[\operatorname{Bernoulli}](#1)}
\NewDocumentCommand{\Dir}{ r() }{\pr[\operatorname{Dir}](#1)}

\IEEEoverridecommandlockouts
\title{\LARGE \bf Learning to combine primitive skills: \\ A step towards versatile robotic manipulation}

\author{Robin Strudel$^{*1}$, Alexander Pashevich$^{*2}$, Igor Kalevatykh$^{1}$, \\
Ivan Laptev$^{1}$, Josef Sivic$^{1}$, Cordelia Schmid$^{2}$}

\begin{document}

\maketitle

\footnotetext[1]{Inria, \'{E}cole normale sup\'{e}rieure, CNRS, PSL Research University, 75005 Paris, France.}
\footnotetext[2]{University Grenoble Alpes, Inria, CNRS, Grenoble INP, LJK, 38000 Grenoble, France.}
\renewcommand*{\thefootnote}{\fnsymbol{footnote}}
\footnotetext[1]{Equal contribution.}
\vspace*{-0.6cm}

\thispagestyle{empty}
\pagestyle{empty}

\begin{abstract}
Manipulation tasks such as preparing a meal or assembling furniture remain highly challenging for robotics and vision. 
Traditional task and motion planning (TAMP) methods can solve complex tasks but require full state observability and are not adapted to dynamic scene changes. 
Recent learning methods can operate directly on visual inputs but typically require many demonstrations and/or task-specific reward engineering.
In this work we aim to overcome previous limitations and propose a reinforcement learning (RL) approach to task planning that learns to combine primitive skills.
First, compared to previous learning methods, 
our approach requires neither intermediate rewards nor complete task demonstrations during training. 
Second, we demonstrate the versatility of our vision-based task planning in challenging settings with temporary occlusions and dynamic scene changes.
Third, we propose an efficient training of basic skills from few synthetic demonstrations by exploring recent CNN architectures and data augmentation. Notably, while all of our policies are learned on visual inputs in simulated environments, we demonstrate the successful transfer and high success rates when applying such policies to manipulation tasks on a real UR5 robotic arm.

\end{abstract}

\section{Introduction}
\vspace{-.1cm}
\label{sec:introduction}

In this work we consider visually guided robotics manipulations and aim to learn robust visuomotor control policies for particular tasks. Autonomous manipulations such as assembling IKEA furniture~\cite{suarez2018can} remain highly challenging given the complexity of real environments as well as partial and uncertain observations provided by the sensors. Successful methods for task and motion planning (TAMP)~\cite{srivastava2014combined,lozano2014constraint,toussaint2015logic} achieve impressive results for complex tasks but often rely on limiting assumptions such as the full state observability and known 3D shape models for manipulated objects. Moreover, TAMP methods usually complete planning before execution and are not robust to dynamic scene changes.

Recent learning methods aim to learn visuomotor control policies directly from image inputs. Imitation learning~(IL) \cite{Pomerleau1989, Ross2014, Pinto2017, Ng2000} is a supervised approach that can be used to learn simple skills from expert demonstrations. One drawback of IL is its difficulty to handle new states that have not been observed during demonstrations. While increasing the number of demonstrations helps to alleviate this issue, an exhaustive sampling of action sequences and scenarios becomes impractical for long and complex tasks.

In contrast, reinforcement learning (RL) requires little supervision and achieves excellent results for some challenging tasks~\cite{Mnih2015Human-levelLearning,Silver2016MasteringSearch}. RL explores previously unseen scenarios and, hence, can generalize beyond expert demonstrations. As full exploration is exponentially hard and becomes impractical for problems with long horizons, RL often relies on careful engineering of rewards designed for specific tasks.

Common tasks such as preparing food or assembling furniture require long sequences of steps composed of many different actions. Such tasks have long horizons and, hence, are difficult to solve by either  RL or IL methods alone. To address this issue, we propose a RL-based method that learns to combine simple imitation-based policies. Our approach  simplifies RL by reducing its exploration to sequences with a limited number of primitive actions, that we call {\em skills}.

\begin{figure*}[t!]
\centering
\mbox{}\vspace{-1.4cm}\\
     \includegraphics[width=0.48\textwidth]{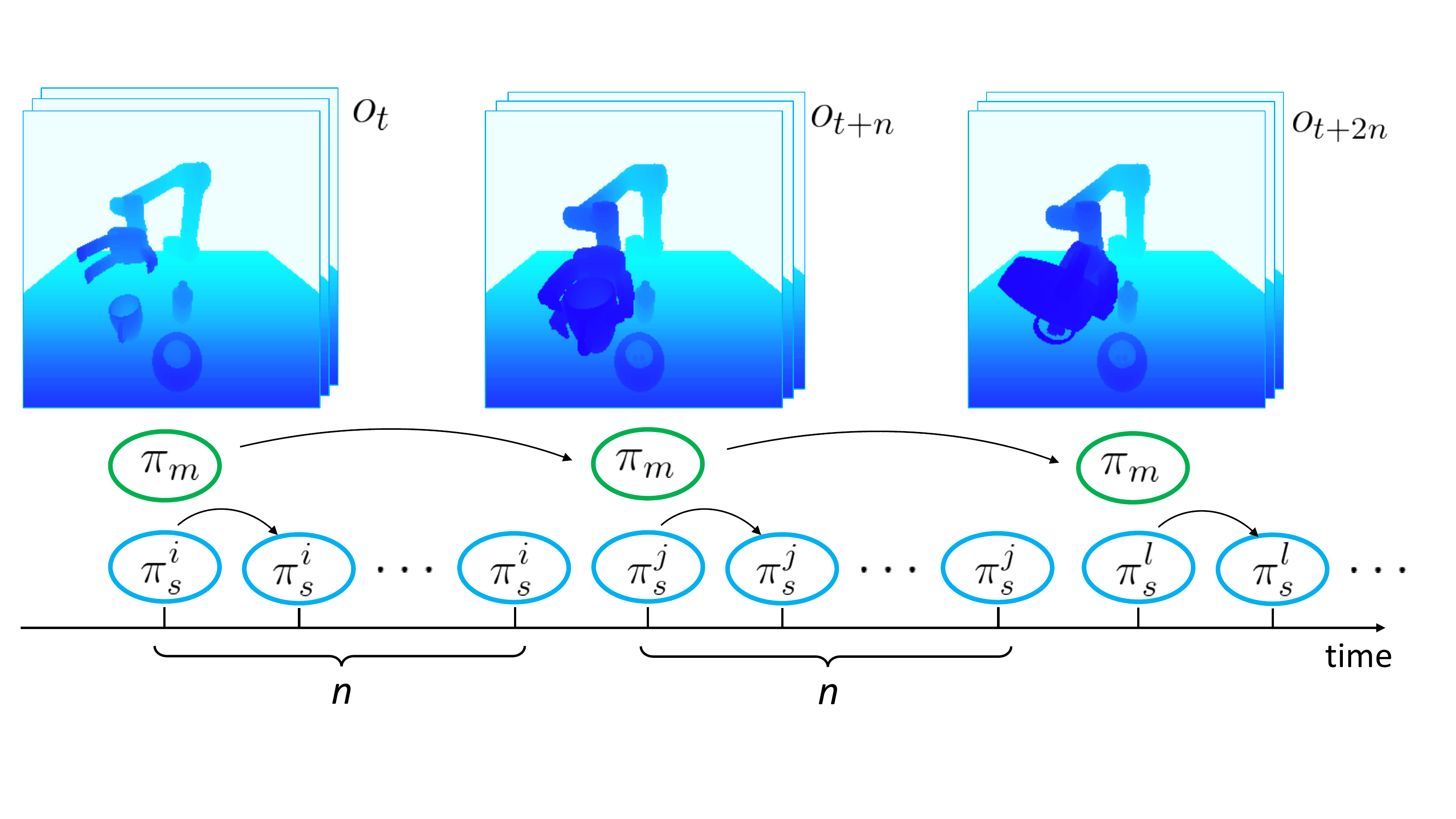}
     \,\,\,\,\,
     \includegraphics[trim=0 -30 100 0, clip, width=0.48\textwidth]{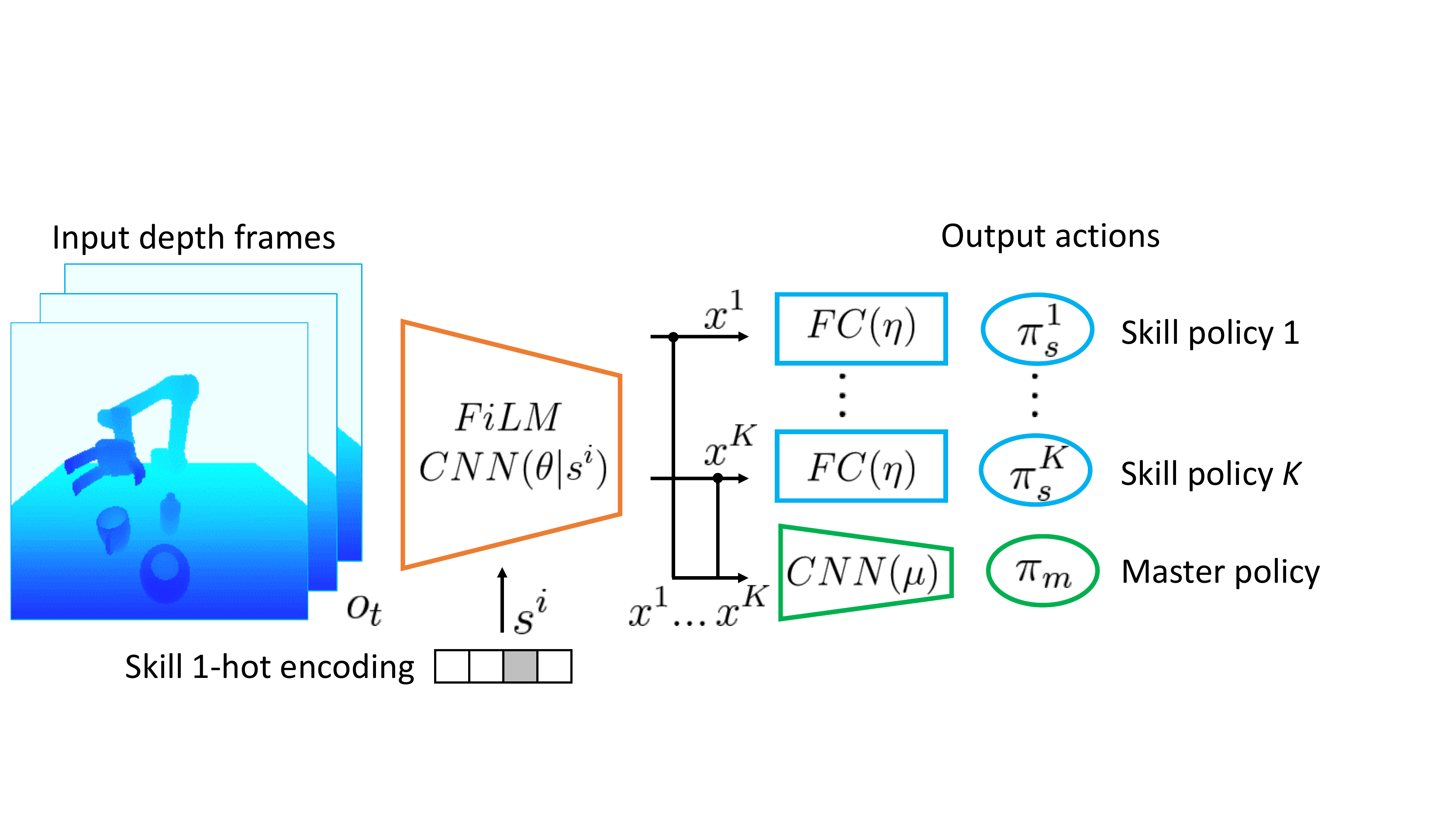}
     \mbox{}\vspace{-.7cm}\\
     \caption{Illustration of our approach. (Left):~Temporal hierarchy of master and skill policies. 
     The master policy $\pi_m$ is executed at a coarse interval of $n$ time-steps to select among $K$ skill policies $\pi_s^1 \ldots \pi_s^K$.
     Each skill policy generates control for a primitive action such as \emph{grasping} or \emph{pouring}.
     (Right):~CNN architecture used for the skill and master policies. }
     \mbox{}\vspace{-0.7cm}\\
     \label{fig:method}
\end{figure*}

Given a set of pre-trained skills such as "grasp a cube" or "pour from a cup", we train RL with sparse binary rewards corresponding to the correct/incorrect execution of the full task. 
While hierarchical policies have been proposed in the past~\cite{Das2018NeuralMC,le2018hieararchical}, our approach can learn {\em composite manipulations using no intermediate rewards and no demonstrations of full tasks}. Hence, the proposed method can be directly applied to learn new tasks. 
See Figure~\ref{fig:method} for an overview of our approach. 

Our skills are low-level visuomotor controllers learned from synthetic demonstrated trajectories with behavioral cloning~(BC)~\cite{Pomerleau1989}.
Examples of skills include {\em go to the bowl}, {\em grasp the object}, {\em pour from the held object}, {\em release the held object}, etc. We automatically generate expert synthetic demonstrations and learn corresponding skills in simulated environments.
We also minimize the number of required demonstrations by choosing appropriate CNN architectures and data augmentation methods.
Our approach is shown to compare favorably to the state of the art~\cite{Pinto2017} on the FetchPickPlace test environment~\cite{Plappert2018}.
Moreover, using recent techniques for domain adaptation~\cite{learningsim2real2019} we demonstrate the successful transfer and high accuracy of our sumulator-trained policies when tested on a real robot

We compare our approach with two classical methods: (a) an open-loop controller estimating object positions and applying a standard motion planner (b) a closed-loop controller adapting the control to re-estimated object positions. We show the robustness of our approach to a variety of perturbations. The perturbations include dynamic change of object positions, new object instances and temporary object occlusions. The versatility of learned policies comes from both the reactivity of the BC learned skills and the ability of the RL master policy to re-plan in case of failure. Our approach allows to compute adaptive control and planning in real-time.

In summary, this work makes the following contributions.
(i)~We propose to learn robust RL policies that combine BC skills to solve composite tasks.
(ii)~We present sample efficient training of BC skills and demonstrate an improvement compared to the state of the art.
(iii)~We demonstrate successful learning of relatively complex manipulation tasks with neither intermediate rewards nor full demonstrations. 
(iv)~We successfully transfer and execute policies learned in simulation to real robot setups.
(v)~We show successful task completion in the presence of perturbations.

Our simulation environments together with the code and models used in this work is publicly available at \url{https://www.di.ens.fr/willow/research/rlbc/}.

\vspace{-.1cm}
\section{Related work}
\vspace{-.1cm}
\label{sec:related_work}

Our work is related to robotics manipulation such as grasping~\cite{Lampe2013AcquiringLearning}, opening doors~\cite{Gu2016}, screwing the cap of a bottle~\cite{Levine2015End-to-EndPolicies} and cube stacking~\cite{Popov2017Data-efficientManipulation}.
Such tasks have been addressed by various methods including imitation learning (IL)~\cite{2017arXiv170307326D} and reinforcement learning (RL)~\cite{SACX}. 

\noindent {\bf Imitation learning (IL).} A neural network is trained to solve a task by observing demonstrations. Approaches include behavioral cloning (BC)~\cite{Pomerleau1989} and inverse reinforcement learning~\cite{Ng2000}. BC learns a function that maps states to expert actions~\cite{Ross2014, Pinto2017}, whereas inverse reinforcement learning learns a reward function from demonstrations in order to solve the task with RL~\cite{Ho2016, Kumar2016LearningImitation, Popov2017Data-efficientManipulation}. BC typically requires a large number of demonstrations and has issues with not observed trajectories. While these problems might be solved with additional expert supervision~\cite{Ross2014} or noise injection in expert demonstrations~\cite{Laskey2017}, we address them by improving the standard BC framework. We use recent state-of-the-art CNN architectures and data augmentation for expert trajectories. This permits to significantly  reduce the number of required demonstrations and to improve performance. 

\noindent {\bf Reinforcement learning (RL).} RL learns to solve a tasks without demonstrations using exploration. Despite impressive results in several  domains~\cite{Silver2016MasteringSearch, Mnih2015Human-levelLearning, Kober2013, Gu2016}, RL methods show limited capabilities when operating in complex and sparse-reward environments common in robotics.
Moreover, RL methods typically require prohibitively large amounts of interactions with the environment during training. 
Hierarchical RL (HRL) methods alleviate some of these problems by learning a high-level policy modulating low-level workers. HRL approaches are generally based either on options~\cite{sutton1999between} or a feudal framework~\cite{Dayan1993FeudalLearning}. The option methods learn a master policy that switches between separate skill policies~\cite{MetaLearningSharedHierarchies, lee2018composing, bacon2017option, Florensa2017StochasticLearning}. The feudal approaches learn a master policy that modulates a low-level policy by a control signal~\cite{LatentSpacePolicies, hiro2018arxiv, Vezhnevets2017FeUdalLearning, kulkarni2016hierarchical, hausman2018learning}. Our approach is based on options but in contrast to the cited methods, we pretrain the skills with IL. This allows us to solve complex and sparse reward problems using significantly less interactions with the environment during training.

\noindent {\bf Combining RL and IL.} A number of approaches combining RL and IL 
have been introduced recently. 
Gao et al.~\cite{gao2018reinforcement} use demonstrations to initialize the RL agent.
\cite{Cheng2018fast, sun2018truncated} use RL to improve expert demonstrations, but do not learn hierarchical policies.
Demonstrations have been also used to define RL objective functions~\cite{Hester2017deepqlearning, Nair2018overcoming}
and rewards~\cite{zhu2018reinforcementAnd}.
Das et al.~\cite{Das2018NeuralMC} combines IL and RL to learn a hierarchical policy. Unlike our method, however, \cite{Das2018NeuralMC} requires full task demonstrations and task-specific reward engineering. Moreover, the addressed navigation problem in~\cite{Das2018NeuralMC} has a much lower time horizon compared to our tasks. \cite{Das2018NeuralMC} also relies on pre-trained CNN representations which limits its application domain.
Le at al.~\cite{le2018hieararchical} train low-level skills with RL, while using demonstrations to switch between skills. 
In a reverse manner, we use IL to learn low-level control and then deploy RL to find appropriate sequences of pre-trained skills. The advantage is that our method can learn a variety of complex manipulations without full task demonstrations.
Moreover, \cite{Das2018NeuralMC,le2018hieararchical} learn discrete actions and cannot be directly applied to robotics manipulations that require continous control.

In summary, none of the methods~\cite{Das2018NeuralMC,le2018hieararchical,Cheng2018fast,sun2018truncated}  is directly suitable for learning complex robotic manipulations due to requirements of dense rewards~\cite{Das2018NeuralMC,sun2018truncated} and state inputs~\cite{Cheng2018fast, sun2018truncated}, limitations to short horizons and discrete actions~\cite{Das2018NeuralMC,le2018hieararchical}, the requirement of full task demonstrations~\cite{Das2018NeuralMC,le2018hieararchical,Cheng2018fast, sun2018truncated} and the lack of learning of visual representations~\cite{Das2018NeuralMC,Cheng2018fast, sun2018truncated}.
Moreover, our skills learned from synthetic demonstrated trajectories outperform RL based methods, see Section~\ref{sec:bc_sota}.

\vspace{-.1cm}

\section{Approach}
\vspace{-.1cm}
\label{sec:approach}

Our RLBC approach aims to learn multi-step policies by combining reinforcement learning (RL) and pre-trained skills obtained with behavioral cloning (BC). We present BC and RLBC in Sections~\ref{sec:bc_method} and \ref{sec:bcrl_method}. Implementation details are given in Section~\ref{sec:implementation_details}.

\vspace{-.2cm}
\subsection{Skill learning with behavioral cloning}
\vspace{-.1cm}
\label{sec:bc_method}

Our first goal is to learn basic skills that can be composed  into more complex policies.
Given observation-action pairs $\mathcal{D}=\{(o_t, a_t)\}$ along expert trajectories, we follow the behavioral cloning approach~\cite{Pomerleau1989} and learn a function approximating the conditional distribution of the expert policy $\pi_\textrm{E}(a_t|o_t)$ controlling a robot arm. 
Our observations $o_t \in \mathcal{O} =\mathbb{R}^{\textrm{H} \times \textrm{W} \times \textrm{M}}$ are sequences of the last $M$ depth frames.
Actions $a_t = (\mathbf{v}_t, \bm{\omega}_t, g_t)$, $a_t \in \mathcal{A}^{\textrm{BC}}$ are defined by the end-effector linear velocity $\mathbf{v}_t \in \mathbb{R}^3$ and angular velocity $\bm{\omega}_t \in \mathbb{R}^3$ as well as the gripper openness state $g_t \in \{0, 1\}$. 

We learn the deterministic skill policies $\pi_s: \mathcal{O} \to \mathcal{A}^{\textrm{BC}}$ approximating the expert policy $\pi_\textrm{E}$. 
Given observations $o_t$ with corresponding expert (ground truth) actions $a_t=(\mathbf{v}_t, \bm{\omega}_t,g_t)$, we represent $\pi_s$ with a convolutional neural network (CNN) and learn network parameters $(\theta, \eta)$ such that predicted actions $\pi_s(o_t) = (\mathbf{\hat{v}}_t, \bm{\hat{\omega}}_t, \hat{g_t})$ minimize the loss
$L_{\textrm{BC}}(\pi_s(o_t), a_t) = \lambda \norm{[\mathbf{\hat{v}_t}, \bm{\hat{\omega}}_t] - [\mathbf{v}_t, \bm{\omega}_t]}^2_2 + (1-\lambda) \left(g_t \log{\hat{g_t}} + (1 - g_t) \log{(1 - \hat{g_t})}\right)$, where $\lambda \in [0, 1]$ is a scaling factor which we empirically set to 0.9.

Our network architecture is presented in Figure~\ref{fig:method}(right).
When training a skill policy $\pi_s^i$, such as reaching, grasping or pouring, we condition the network on the skill using the recent FiLM architecture~\cite{perez2018film}. Given the one-hot encoding $s^i$ of a skill $i$, we use $s^i$ as input to the FiLM generator which performs affine transformations of the network feature maps. FiLM conditions the network on performing a given skill, which permits learning a shared representation for all skills. Given an observation $o_t$, the network $CNN(\theta|s^i)$ generates a feature map $x^i_t$ conditioned on skill $i$. The spatially-averaged $x_i$ is linearly mapped with $FC(\eta)$ to the action of $\pi_s^i$.

\vspace{-.2cm}
\subsection{RLBC approach}
\vspace{-.1cm}
\label{sec:bcrl_method}
We wish to solve composite manipulations without full expert demonstrations and with a single sparse reward.
For this purpose we rely on a high-level {\em master policy} $\pi_m$ controlling the pre-trained {\em skill policies} $\pi_s$ at a coarse timescale.
To learn $\pi_m$, we follow the standard formulation of reinforcement learning and
maximize the expected return $\mathbb{E}_{\pi} \sum_{k=0}^{\infty} \gamma^k r_{t+k}$ 
given rewards $r_{t}$. 
Our reward function is sparse and returns $1$ upon successful termination of the task and $0$ otherwise. The RL master policy $\pi_m: \mathcal{O} \times \mathcal{A}^{\textrm{RL}} \to [0, 1]$ chooses one of the $K$ skill policies to execute the low-level control, i.e., the action space of $\pi_m$ is discrete: $\mathcal{A}^{\textrm{RL}} = \{1, \ldots, K\}$. 
Note, that our sparse reward function makes the learning of deep visual representations challenging. We, therefore, train $\pi_m$ using visual features $x_t^i$ obtained from the BC pre-trained $CNN(\theta|s^i)$.
Given an observation $o_t$, we use the concatenation of skill-conditioned features $\{x^1_t, \ldots, x^K_t\}$ as input for the master $CNN(\mu)$, see Figure~\ref{fig:method}(right).

To solve composite tasks with sparse rewards, we use a coarse timescale for the master policy. 
The selected skill policy controls the robot for $n$ consecutive time-steps before the master policy is activated again to choose a new skill. 
This allows the master to focus on high-level task planning rather than low-level motion planning achieved by the skills. We expect the master policy to recover from unexpected events, for example, if an object slips out of a gripper, by re-activating an appropriate skill policy.
Our combination of the master and skill policies is illustrated in Figure~\ref{fig:method}(left).

\textit{RLBC algorithm.} The pseudo-code for the proposed approach is shown in Algorithm~\ref{alg:bcrl}. The algorithm can be divided into three main steps. 
First, we collect a dataset of expert trajectories $\mathcal{D}_k$ for each skill policy $\pi^k_s$.  For each policy, we use an expert script that has an access to the full state of the environment.
Next, we train a set of skill policies $\{\pi^1_s, \ldots, \pi^K_s\}$. We sample a batch of state-action pairs and update parameters of convolutional layers $\theta$ and the skills linear layer parameters $\eta$.
Finally, we learn the master $\pi_m$ using the pretrained skill policies and the frozen parameters $\theta$. We collect episode rollouts by first choosing a skill policy with the master and then applying the selected skill to the environment for $n$ time-steps.
We update the master policy weights $\mu$ to maximize the expected sum of rewards.

\vspace{-0.2cm}
\begin{algorithm}
\caption{RLBC}
\label{alg:bcrl}
\begin{algorithmic}[1]
\begin{footnotesize}
\State *** Collect expert data ***
\For{$k \in \{1, \ldots, K\}$}
    \State Collect an expert dataset $\mathcal{D}_k$ for the skill policy $\pi^k_s$
\EndFor
\State *** Train $\{\pi^1_s, \ldots, \pi^K_s\}$ by solving: ***
\State $\theta, \eta = \arg\min_{\theta, \eta} \sum_{k=1}^{K} \sum_{(o_t, a_t) \in \mathcal{D}_k} L_{\textrm{BC}}(\pi^k_s(o_t), a_t)$
\While{task is not solved}
    \State *** Collect data for the master policy ***
    \State $\mathcal{E} = \{\}$ \Comment{Empty storage for rollouts}
    \For{$\textrm{episode\_id} \in \{1, \ldots, \textrm{ppo\_num\_episodes}\}$}
        \State $o_0 = \textrm{new\_episode\_observation}()$
        \State $t = 0$
        \While{episode is not terminated}
            \State $k_{t} \sim \pi_m(o_t)$ \Comment{Choose the skill policy}
            \State $o_{t+n}, r_{t+n} = \textrm{perform\_skill}(\pi^{k_{t}}_s, o_t)$
            \State $t = t + n$
        \EndWhile
        \State $\mathcal{E} = \mathcal{E} \cup \{(o_0, k_1, r_1, o_1, k_2, r_2, o_2, \ldots)\}$
    \EndFor
    \State *** Make a PPO step for the master policy on $\mathcal{E}$ ***
    
    \State $\mu = \textrm{ppo\_update}(\pi_m, \mathcal{E})$
\EndWhile
\mbox{}\vspace{-0.5cm}\\
\end{footnotesize}
\end{algorithmic}
\end{algorithm}

\subsection{Approach details}
\label{sec:implementation_details}

\textit{Skill learning with BC.} 
We use ResNet-18 for the $CNN(\theta|s^i)$, which we compare to VGG16 and ResNet-101 in Section~\ref{sec:bc_networks}.
We augment input depth frames with random translations, rotations and crops. We also perform viewpoint augmentation and sample the camera positions on a section of a sphere centered on the robot and with a radius of $1.40\,\text{m}$. We uniformly sample the yaw angle in $[-15\degree, 15\degree]$, the pitch angle in $[15\degree, 30\degree]$, and the distance to the robot base in $[1.35, 1.50]$ m. The impact of both augmentations is evaluated in Section~\ref{sec:bc_viewpoints}.
We normalize the ground truth of the expert actions to have zero mean and a unit variance and normalize the depth values of input frames to $[-1, 1]$.
We learn BC skills using Adam~\cite{Diederik2014Adam} with the learning rate $10^{-3}$ and a batch size 64.
We also use Batch Normalization~\cite{Ioffe2015BatchShift}.

\textit{Task learning with RL.} 
We learn the master policies with the PPO~\cite{PPO} algorithm using the open-source implementation~\cite{pytorchrl} where we set the entropy coefficient to 0.05, the value loss coefficient to 1, and use 8 episode rollouts for the PPO update.
For the RLBC method, the concatenated skill features $\{x_t^1,\ldots,x_t^K\}$ are processed with the master network $CNN(\mu)$ having 2 convolutional layers with 64 filters of size $3 \times 3$.
During pre-training of skill policies we update the parameters $(\theta, \eta)$. When training the master policy, we only update $\mu$ while keeping ($\theta$, $\eta$) parameters fixed.
We train RLBC using 8 different random seeds in parallel and evaluate the best one.

\textit{Real robot transfer.} To apply our method on the real robot, we use a state-of-the-art technique of learning sim2real transfer based on data augmentation with domain randomization~\cite{learningsim2real2019}. This method uses a proxy task of cube position prediction and a set of basic image transformations to learn a sim2real data augmentation function for depth images. We augment the depth frames from synthetic expert demonstrations with this method and, then, train skill policies. Once the skill policy is trained on these augmented simulation images, it is directly used on the real robot.

\vspace{-.1cm}

\section{Experimental setup}
\vspace{-.1cm}
\label{sec:setup}
This section describes the setup used to evaluate our approach.  First, we present the robot environment and the different tasks in Sections~\ref{sec:setup_robot} and \ref{sec:setup_tasks}. Next, we describe the synthetic dataset generation and skill definition for each task in Sections~\ref{sec:setup_datasets} and \ref{sec:setup_skills}.

\vspace{-.2cm}
\subsection{Robot and agent environment}
\begin{figure*}
\begin{subfigure}{.31\textwidth}
  \centering
   \includegraphics[width=0.46\textwidth, trim={0 0.8cm 0 0.4cm}, clip]{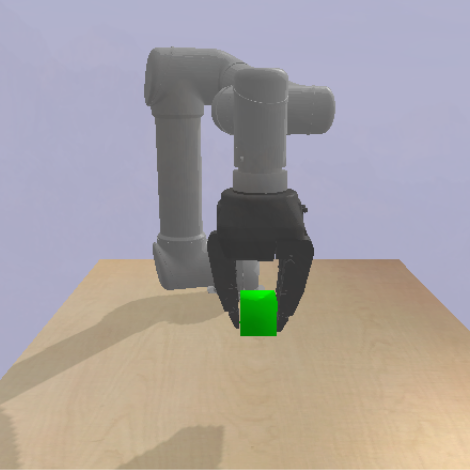}
   \includegraphics[width=0.46\textwidth, trim={0 0.8cm 0 0.4cm}, clip]{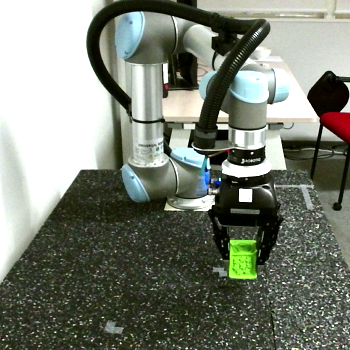}
  \caption{\footnotesize UR5-Pick}
  \label{fig:env_pick}
\end{subfigure} \,\
\begin{subfigure}{.31\textwidth}
  \centering
   \includegraphics[width=0.46\textwidth, trim={0 0.8cm 0 0.4cm}, clip]{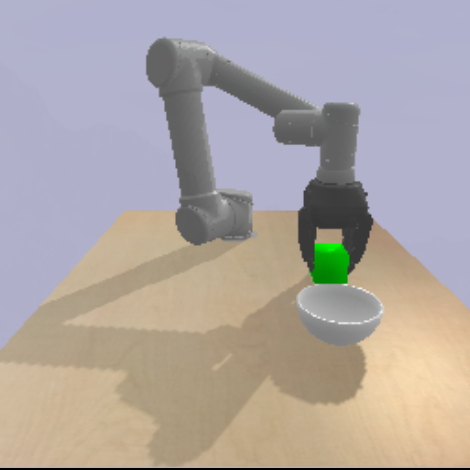}
   \includegraphics[width=0.46\textwidth, trim={0 0.8cm 0 0.4cm}, clip]{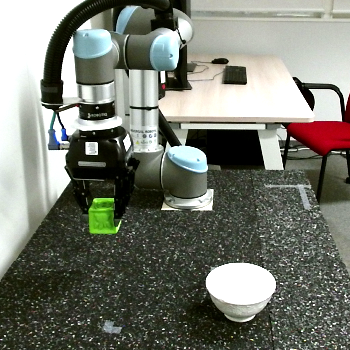}
  \caption{\footnotesize UR5-Bowl}
  \label{fig:env_bowl}
\end{subfigure} \,\
\begin{subfigure}{.31\textwidth}
  \centering
   \includegraphics[width=0.46\textwidth, trim={0 0.2cm 0 1.0cm}, clip]{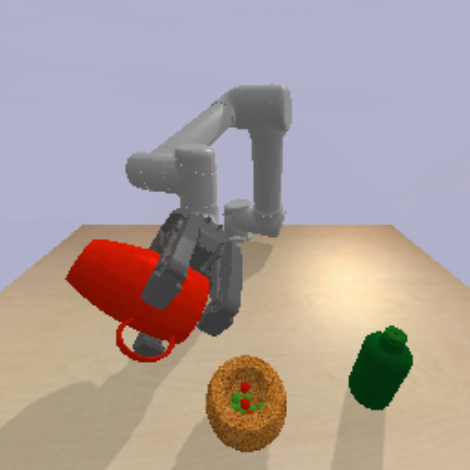}
   \includegraphics[width=0.46\textwidth, trim={0 0.2cm 0 1.0cm}, clip]{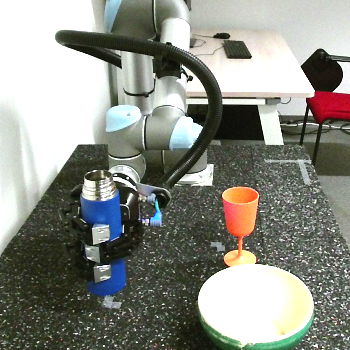}
  \caption{\footnotesize UR5-Breakfast}
  \label{fig:env_breakfast}
\end{subfigure}\\[-0.2cm]
  \caption{UR5 tasks used for evaluation: (a) task of picking up the cube, (b) task of bringing the cube to the bowl, (c) task of pouring the cup and the bottle into the bowl. (Left) simulation, (right) real robot.}
 \vspace{-0.7cm}
\label{fig:envs}
\end{figure*}

\vspace{-.1cm}
\label{sec:setup_robot}
For our experiments we use a 6-DoF UR5 robotic arm with a 3 finger Robotiq gripper, see Figure~\ref{fig:envs}. In simulation, we model the robot with the \texttt{pybullet} physics simulator~\cite{Courmans2016}.
For observation, we record depth images with the Microsoft Kinect 2 placed in front of the arm. 
The agent takes as input the three last depth frames $o_t \in \mathbb{R}^{224 \times 224 \times 3}$ and commands the robot with an action $a_t \in \mathbb{R}^7$. The control is performed at 10 Hz frequency.

\vspace{-.2cm}
\subsection{UR5 tasks}
\vspace{-.1cm}
\label{sec:setup_tasks}

For evaluation, we consider 3 tasks: UR5-Pick, UR5-Bowl and UR5-Breakfast.
The {\bf UR5-Pick} task  picks up a cube of a size between 3.5 cm and 8.0 cm and lifts it up, see Figure~\ref{fig:env_pick}.
In {\bf UR5-Bowl} the robot has to grasp the cube and place it in the bowl, see Figure~\ref{fig:env_bowl}.
The {\bf UR5-Breakfast} task contains a cup, a bottle and a bowl as shown in Figure~\ref{fig:env_breakfast}.
We use distinct ShapeNet~\cite{Chang2015} object instances for the training and test sets (27 bottles, 17 cups, 32 bowls in each set).
The robot needs to pour ingredients from the cup and the bottle in the bowl. In all tasks, the reward is positive if and only if the task goal is reached. The maximum episode lengths are 200, 600, and 2000 time-steps for UR5-Pick, UR5-Bowl, and UR5-Breakfast correspondingly.

\vspace{-.2cm}
\subsection{Synthetic datasets}
\vspace{-.1cm}
\label{sec:setup_datasets}

We use the simulated environments to create a synthetic training and test set. For all our experiments, we collect trajectories with random initial configurations where the objects and the end-effector are allocated within a workspace of $80\times 40\times 20\, \text{cm}^3$. The synthetic demonstrations are collected using an expert script designed for each skill. The script has access to the full state of the system including the states of the robot and the objects. To generate synthetic demonstrations, we program end-effector trajectories and use inverse kinematics (IK) to generate corresponding trajectories in the robot joints space. 
Each demonstration consists of multiple pairs of the three last camera observations and the robot control command performed by the expert script.
For UR5-Pick, we collect $1000$ synthetic demonstrated trajectories for training.
For UR5-Bowl and UR5-Breakfast, we collect a training dataset of 250 synthetic demonstrations.
For evaluation of each task, we use 100 different initial configurations in simulation and 20 trials on the real robot.

\vspace{-.2cm}
\subsection{Skill definition}
\vspace{-.1cm}
\label{sec:setup_skills}

UR5-Pick task is defined as a single skill. For UR5-Bowl and UR5-Breakfast, we consider a set of skills defined by expert scripts. For UR5-Bowl, we define four skills: (a)~go to the the cube, (b)~go down and grasp, (c)~go up, and (d)~go to the bowl and open the gripper. 
For UR5-Breakfast, we define four skills: (a)~go to the bottle, (b)~go to the cup, (c) grasp an object and pour it to the bowl, and (d)~release the held object.
We emphasize that the expert dataset does not contain full task demonstrations and that all our training is done in simulation.
When training the RL master, we execute selected skills for 60 consecutive time-steps for the UR5-Bowl task and 220 time-steps for the UR5-Breakfast task.

\vspace{-.1cm}
\section{Evaluation of BC skill learning}
\vspace{-.1cm}
\label{sec:bc_evaluation}

This section evaluates the different parameters of the BC skill training for the UR5-Pick task and a comparison with the state of the art. First, we evaluate the impact of the CNN architecture and data augmentation on the skill performance in Sections~\ref{sec:bc_networks} and \ref{sec:bc_viewpoints}. Then, we show that the learned policies transfer to a real-robot in Section~\ref{sec:bc_real}. Finally, we compare the BC skills with the state of the art in Section~\ref{sec:bc_sota}.

\vspace{-.2cm}
\subsection{CNN architecture for BC skill learning}
\vspace{-.1cm}
\label{sec:bc_networks}

\begin{table}
\small
\setlength\tabcolsep{2.7pt} 
\centering
\mbox{}\vspace{.2cm}\\
\begin{tabular}{cccc}
\toprule
Demos & VGG16-BN & ResNet-18 & ResNet-101 \\
\midrule
20 & 1\%      & 1\%       & 0\%        \\
50 & 9\%      & 5\%       & 5\%        \\
100 & 37\%     & 65\%      & 86\%       \\
1000 & 95\%     & \textbf{100\%}     & \textbf{100\%} \\
\bottomrule
\end{tabular}

\caption{Evaluation of BC skills trained with different CNN architectures and number of demonstrations on the UR5-Pick task in simulation.}
\label{tab:bc_archis}
\end{table}
\begin{table}
\small
\setlength\tabcolsep{2.7pt} 
\centering
\mbox{}\vspace{-.3cm}\\
\begin{tabular}{ccccc}
\toprule
\
Demos & None & Standard & Viewpoint & \begin{tabular}[c]{@{}c@{}}Standard \& \\ Viewpoint\end{tabular} \\
\midrule
20 & 1\% & 49\% & 39\% & 75\% \\
50 & 5\% & 81\% & 79\% & 93\% \\
100 & 65\% & 97\% & \textbf{100\%} & \textbf{100\%} \\
\bottomrule
\end{tabular}

\caption{Evaluation of ResNet-18 BC skills trained with different data augmentations on UR5-Pick task in simulation.}
\mbox{}\vspace{-1.0cm}\\
\label{tab:bc_aug}
\end{table}

Given the simulated UR5-Pick task illustrated in Figure~\ref{fig:env_pick}(left), we compare BC skill networks trained with different CNN architectures and varying number of expert demonstrations.
Table~\ref{tab:bc_archis} compares the success rates of policies with VGG and ResNet architectures.
Policies based on the VGG architecture~\cite{Simonyan2014} obtain success rate below $40\%$ with  $100$ training demonstrations and reach $95\%$ with $1000$ demonstrations. 
ResNet~\cite{He2016DeepRecognition} based policies have a success rate above $60\%$ when trained on a dataset of $100$ demonstrations and reach $100\%$ with $1000$ demonstrations. Overall ResNet-101 has the best performance closely followed by ResNet-18 and outperforms VGG significantly. To conclude, we find that the network architecture has a fundamental impact on the BC performance. 
In the following experiments we use ResNet-18 as it presents a good trade-off between performance and training time.

When examining why VGG-based BC has a lower success rate, we observe that it has higher validation errors compared to ResNet. This indicates that VGG performs worse on the level of individual steps and is hence expected to result in higher compounding errors.

\vspace{-.2cm}
\subsection{Evaluation of data augmentation}
\vspace{-.1cm}
\label{sec:bc_viewpoints}

We evaluate the impact of different types of data augmentations in Table~\ref{tab:bc_aug}.
We compare training without data augmentation with 3 variants: (1) random translations, rotations and crops, as is standard for object detection, (2) record each expert synthetic demonstration from 10 varying viewpoints and (3) the combination of  (1) and (2).

Success rates for UR5-Pick on datasets with 20, 50 and 100 demonstrations are reported in Table~\ref{tab:bc_aug}.
We observe that data augmentation is particularly important when only a few demonstrations are available. For 20 demonstrations, the policy trained with no augmentation performs at 1\% while the policy trained with standard and viewpoint augmentations together performs at 75\%.
The policy trained with a combination of both augmentation types performs the best and achieves 93\% and 100\% success rate for  50 and 100 demonstrations respectively. 
In summary, data augmentation allows a significant reduction in the number of expert trajectories required to solve the task.

\vspace{-.2cm}
\subsection{Real robot experiments}
\vspace{-.1cm}
\label{sec:bc_real}

We evaluate our method on the real-world UR5-Pick illustrated in Figure~\ref{fig:env_pick}(right).
We collect demonstrated trajectories in simulation and train the BC skills network applying standard, viewpoint and sim2real augmentations.
We show that our approach transfers well to the real robot using no real images. The learned policy manages to pick up cubes of 3 different sizes correctly in 20 out of 20 trials.

\vspace{-.2cm}
\subsection{Comparison with state-of-the-art methods}
\vspace{-.1cm}
\label{sec:bc_sota}

\begin{figure}
\centering
\begin{subfigure}{.16\textwidth}
  \centering
  \includegraphics[width=\linewidth, height=2.9cm, trim={5cm 8cm 4.5cm 1.5cm}, clip]{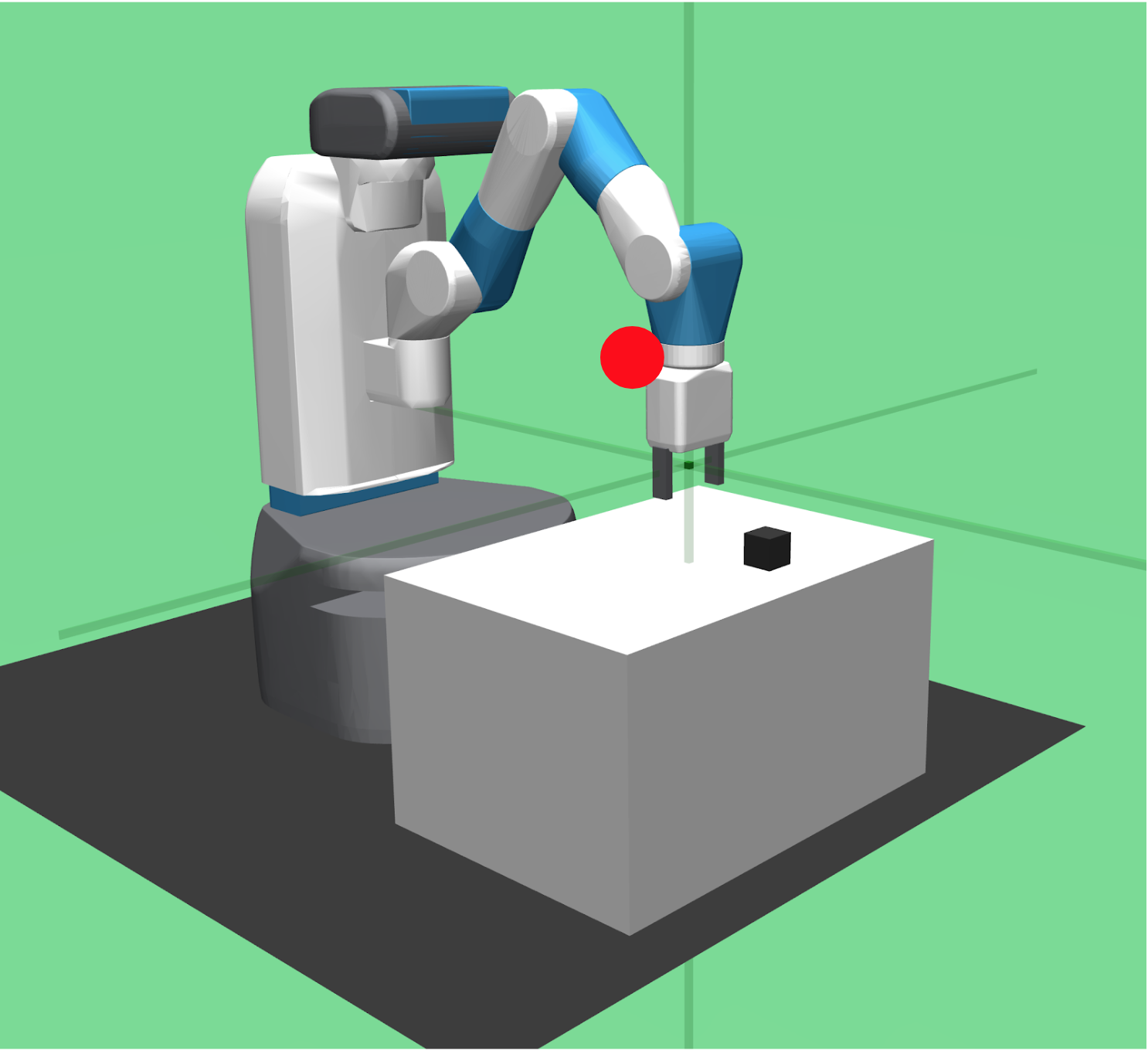}
  \caption{\footnotesize FetchPickPlace}
  \label{fig:fetchpickplace}
\end{subfigure}%
\begin{subfigure}{.26\textwidth}
  \centering
  \includegraphics[width=\linewidth, height=2.9cm, trim={0 0.8cm 0 0.3cm}, clip]{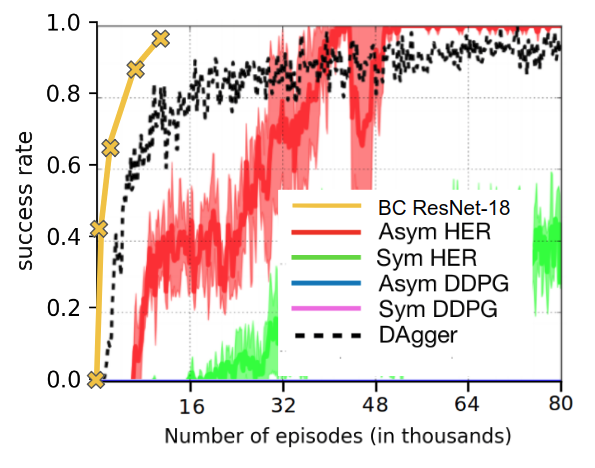}
  \caption{\footnotesize Comparison with \cite{Pinto2017}}
  \label{fig:bc_sota_plot}
\end{subfigure}\\[-0.2cm]
\caption{Comparison of BC ResNet-18 with state of the art~\cite{Pinto2017} on the FetchPickPlace task. BC ResNet-18 results are reported for 200 different initial configurations.}
\vspace{-0.7cm}
\end{figure}
   
One of the few test-beds for robotic manipulation is FetchPickPlace  from OpenAI Gym~\cite{Plappert2018} implemented in \texttt{mujoco}~\cite{Todorov2012MuJoCo:Control},
see Figure~\ref{fig:fetchpickplace}. The goal for the agent is to pick up a cube and to move it to the red target (see Figure~\ref{fig:fetchpickplace}). The agent observes the three last RGB-D images from a camera placed in front of the robot $o_t \in \mathbb{R}^{100 \times 100 \times 4 \times 3}$. The positions of the cube and the target are set at random for each trial. The reward of the task is a single sparse reward of success. The maximum length of the task 
is 50 time-steps. 

For a fair comparison with~\cite{Pinto2017}, we do not use any data augmentation. We report the success rate of ResNet-18 policy in Figure~\ref{fig:bc_sota_plot}. We follow~\cite{Pinto2017} and plot the success rate of both RL and IL methods with respect to the number of episodes used (either trial episodes or 
demonstrations). Our approach outperforms the policies trained with an imitation learning method DAgger~\cite{Ross2014} in terms of performance and RL methods such as HER~\cite{Andrychowicz2017HindsightER} and DDPG~\cite{Lillicrap2016ContinuousLearning} in terms of data-efficiency.
According to~\cite{Pinto2017}, DAgger does not reach 100\% even after $8*10^4$ demonstrations despite the fact that it requires an expert during training. HER reaches the success rate of 100\% but requires about $4* 10^4$ trial episodes. Our approach achieves the 96\% success rate using $10^4$ demonstrations.

Our policies differ from~\cite{Pinto2017} mainly in the CNN architecture. Pinto et al.~\cite{Pinto2017} use a simple CNN with 4 convolutional layers while we use ResNet-18. Results of this section confirm the large impact of the CNN architecture on the performance of visual BC policies, as was already observed in Table~\ref{tab:bc_archis}.

\vspace{-.1cm}

\vspace{-.1cm}
\section{Evaluation of RLBC}
\vspace{-.1cm}
\label{sec:bcrl_evaluation}

This section evaluates the proposed RLBC approach and compares it to baselines introduced in Section~\ref{sec:bcrl_baselines}. 
First, we evaluate our method on UR5-Bowl in Section~\ref{sec:setup_tasks}. 
We then test the robustness of our approach to various perturbations  such as dynamic changes of object positions, dynamic occlusions, unseen object instances and the increased probability of collisions due to small distances between objects. We show that RLBC outperforms the baselines on those scenarios both in simulation and on a real robot.
Note, that our real robot experiments are performed with skills and master
policies that have been trained exclusively in simulation using sim2real
augmentation~\cite{learningsim2real2019}. We use the same policies for all perturbation scenarios. Qualitative results of our method are available in the {\bfseries supplementary video}.

\vspace{-.2cm}
\subsection{Baseline methods}
\vspace{-.1cm}
\label{sec:bcrl_baselines}

We compare RLBC with 3 baselines: (a) a fixed sequence of BC skills following the manually pre-defined correct order (BC-ordered); (b) an open-loop controller estimating positions of objects and executing an expert script (Detect \& Plan); (c) a closed-loop controller performing the same estimation-based control and replanning in case if object positions change substantially (Detect \& Replan).
We use the same set of skills for RLBC and BC-ordered. We train the position estimation network using a dataset of 20.000 synthetic depth images with randomized object positions. All networks use ResNet-18 architecture and are trained with the standard, viewpoint and sim2real augmentations described in Section~\ref{sec:bc_viewpoints}.

\vspace{-.2cm}
\subsection{Results on UR5-Bowl with no perturbations}
\vspace{-.1cm}
\label{sec:bcrl_results_bowl}

\begin{table}
\small
\setlength\tabcolsep{2.7pt} 
\centering
\begin{tabular}{ccccc}
\toprule
\
\begin{tabular}[c]{@{}c@{}}UR5-Bowl \\ perturbations\end{tabular} & \begin{tabular}[c]{@{}c@{}}Detect \& \\ Plan\end{tabular} & \begin{tabular}[c]{@{}c@{}}Detect \& \\ Replan\end{tabular} & BC-ordered & RLBC \\
\midrule
No perturbations & 17/20 & 16/20 & 17/20 & \textbf{20/20} \\
Moving objects & 0/20 & 12/20 & 13/20 & \textbf{20/20} \\
Occlusions & 17/20 & 10/20 & 2/20 & \textbf{18/20} \\
New objects & 16/20 & 14/20 & 15/20 & \textbf{18/20} \\
\bottomrule
\mbox{}\vspace{-.5cm}\\
\end{tabular}
\caption{Comparison of RLBC with 3 baselines on the real-world UR5-Bowl task with dynamic changes of the cube position, dynamic occlusions and new object instances.}
\mbox{}\vspace{-.6cm}\\
\label{tab:hrlbc_bowl_ablation_real}
\end{table}

\begin{figure}
\captionsetup[subfigure]{justification=centering}
\begin{subfigure}{.25\textwidth}
  \centering
  \includegraphics[width=\linewidth]{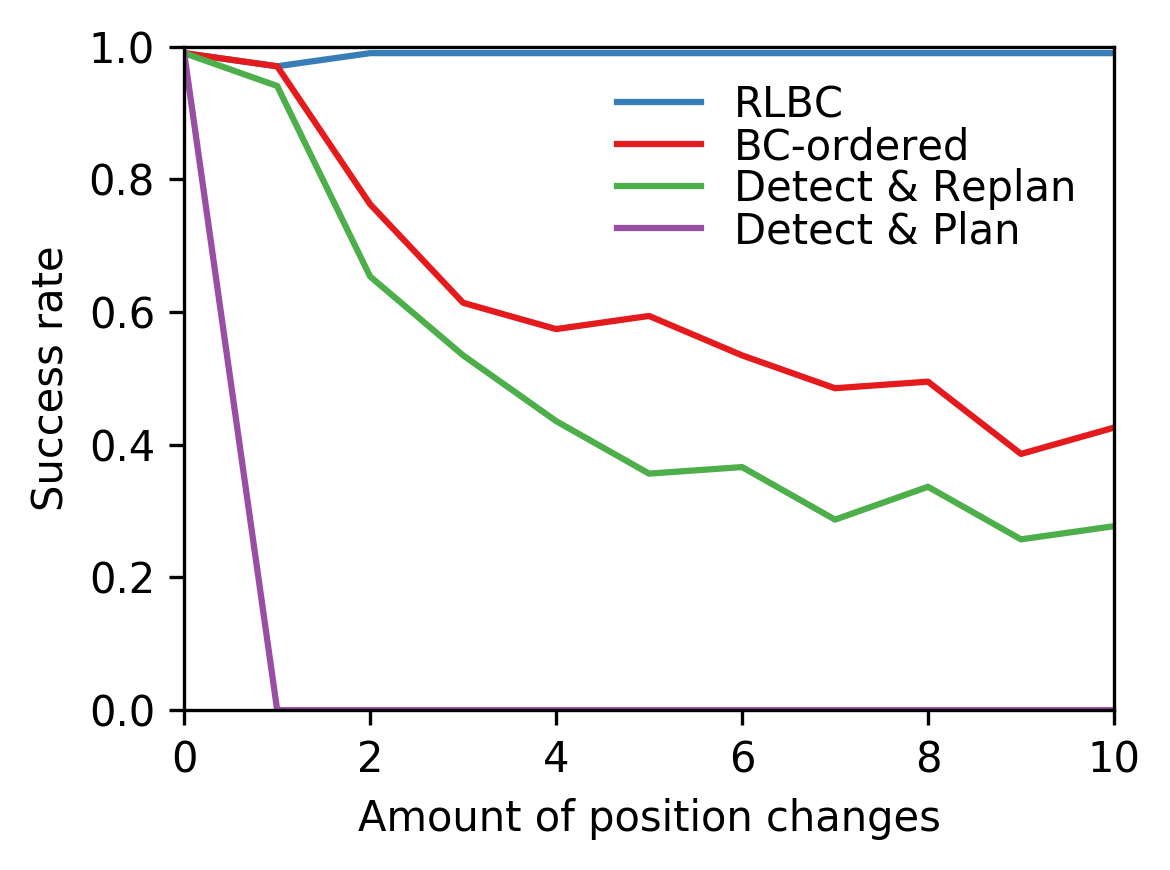}
  \vspace{-.7cm}
  \caption{\footnotesize UR5-Bowl}
  \label{fig:bcrl_bowl_moving}
\end{subfigure}%
\begin{subfigure}{.25\textwidth}
  \centering
  \includegraphics[width=\linewidth]{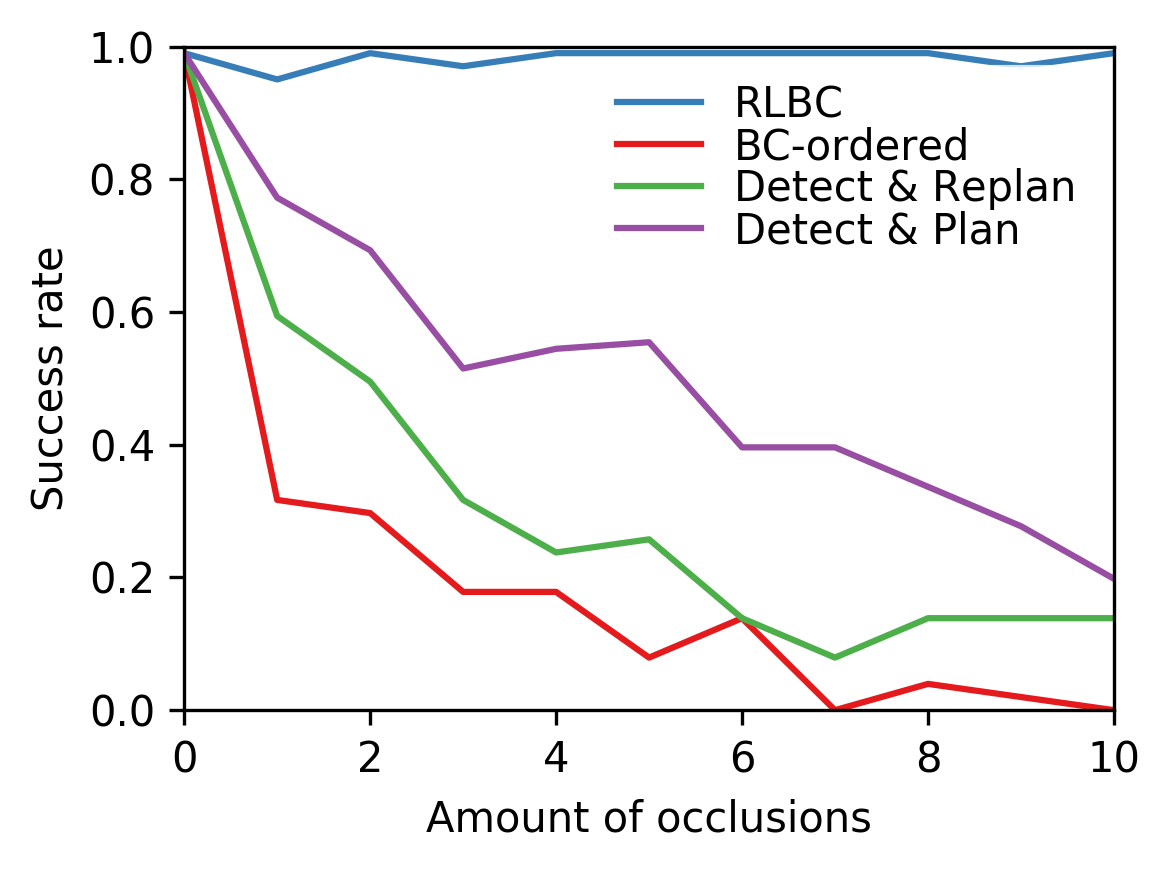}
  \vspace{-.7cm}
  \caption{\footnotesize UR5-Bowl}
  \label{fig:bcrl_bowl_occlusion}
\end{subfigure}\\
\begin{subfigure}{.25\textwidth}
  \centering
  \includegraphics[width=\linewidth]{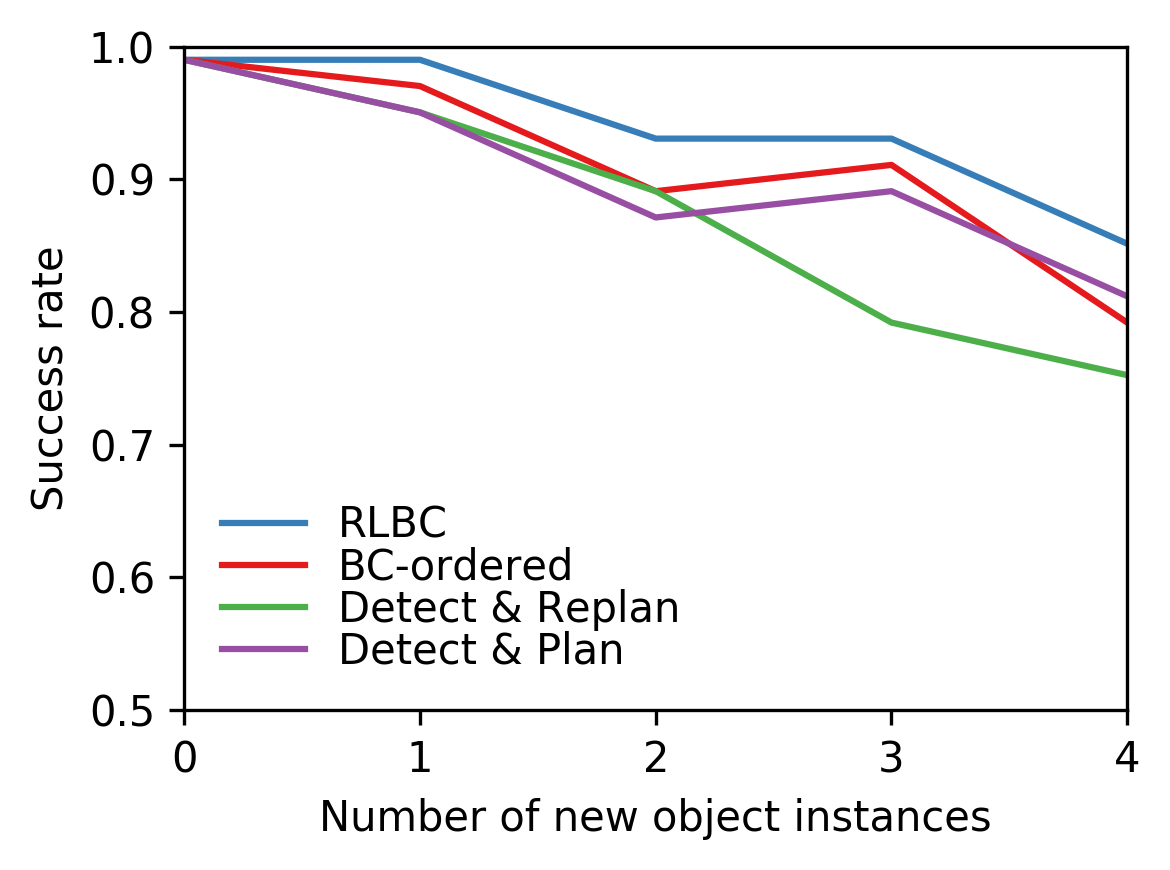}
  \vspace{-.7cm}
  \caption{\footnotesize UR5-Bowl}
  \label{fig:bcrl_bowl_unseen}
\end{subfigure}%
\begin{subfigure}{.25\textwidth}
  \centering
  \includegraphics[width=\linewidth]{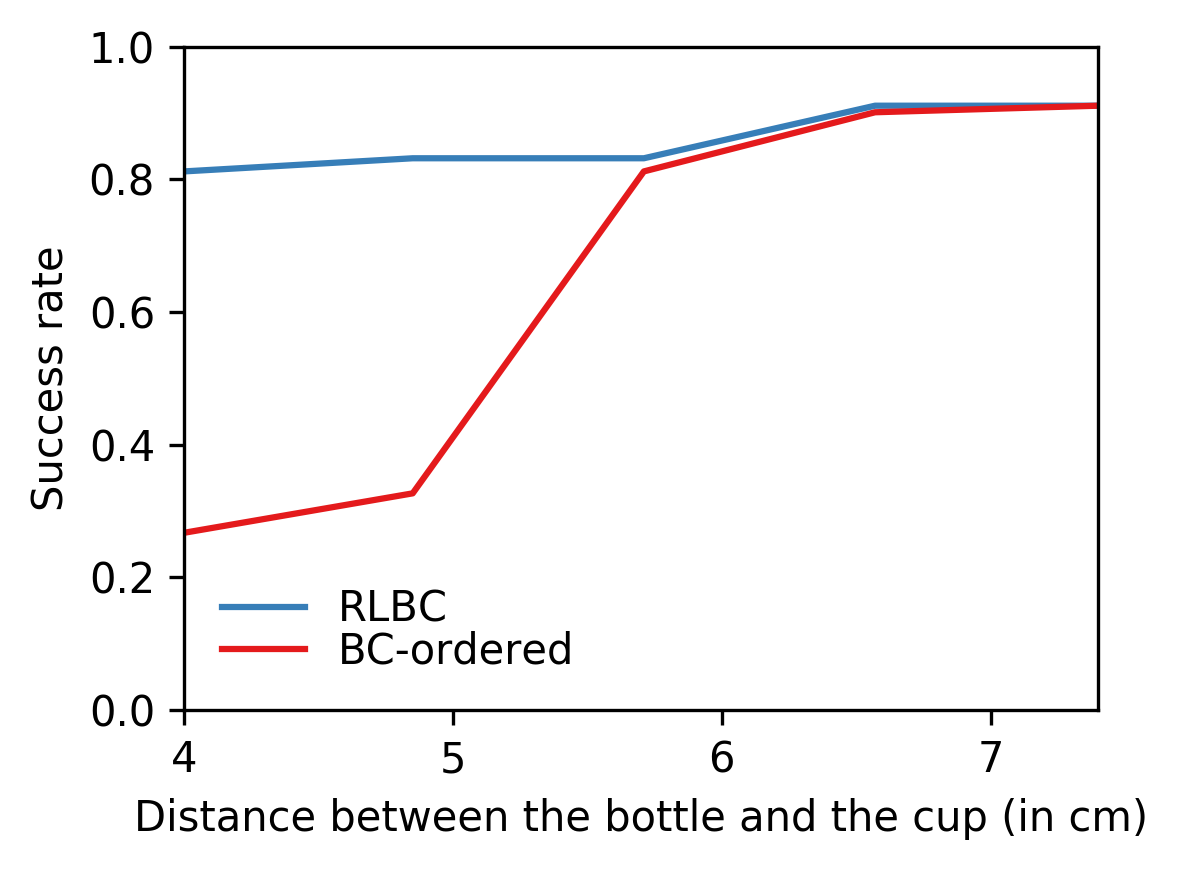}
  \vspace{-.7cm}
  \caption{\footnotesize UR5-Breakfast}
  \label{fig:bcrl_breakfast_distance}
\end{subfigure}
    \caption{Performance of RLBC and baseline methods in simulated environments under perturbations: (a) Dynamic changes of cube position; (b) Dynamic occlusions; (c) Replacing the cube by unseen objects; (d) Decreasing the distance between objects.}
    \vspace{-0.7cm}
\label{fig:bcrl_ablation_sim}
\end{figure}


We first evaluate RLBC and the three baselines on the UR5-Bowl task (see Figure~\ref{fig:env_bowl}).
When tested in simulation, all the baselines and RLBC manage to perfectly solve the task. 
On the real-world UR5-Bowl task, BC-ordered and Detect \& Plan baselines sometimes fail to grasp the object which leads to task failures (see Table~\ref{tab:hrlbc_bowl_ablation_real}, first row). On the contrary, RLBC solves the task in all 20 episodes given its ability to re-plan the task in the cases of failed skills.

We have also attempted to solve the simulated UR5-Bowl task without skills by learning an RL policy performing low-level control. We have used ImageNet pre-trained ResNet-18 to generate visual features. The features were then used to train low-level RL control policy with PPO. Whereas such a low-level RL policy did not solve the task a single time after $10^4$ episodes, RLBC reaches 100\% after 400 episodes.

\vspace{-.2cm}
\subsection{Robustness to perturbations}
\vspace{-.1cm}

\noindent {\it Robustness to dynamic changes in object position.}
We evaluate RLBC against the baselines in the UR5-Bowl scenario where the cube is moved several times during the episode. We plot success rates evaluated in simulation with respect to the number of position changes in Figure~\ref{fig:bcrl_bowl_moving}.
We observe the stability of RLBC and the fast degradation of all baselines.
As both RLBC and BC-ordered use the same set of skills, the stability of RLBC comes from the learned skill combination.
The "Moving objects" row in Table~\ref{tab:hrlbc_bowl_ablation_real} reports results for 3 moves of the cube evaluated on the real robot. Similar to simulated results, we observe excellent results of RLBC and the degraded performance for all the baselines.

\noindent{\it Robustness to occlusions.}
We evaluate the success of UR5-Bowl task under occlusions. Each occlusion lasts 3 seconds and covers a large random part of the workspace by a cardboard. Figure~\ref{fig:bcrl_bowl_occlusion} shows success rates with respect to the number of occlusions in the simulated UR5-Bowl environment. Similar to perturbation results in Figure~\ref{fig:bcrl_bowl_moving}, RLBC demonstrates high robustness to occlusions while the performance of other methods quickly degrades. 
The "Occlusions" row in Table~\ref{tab:hrlbc_bowl_ablation_real} reports results for a single occlusion performed during the real-robot evaluation.
Baseline methods are strongly influenced by occlusions except Detect \& Plan which performs well unless occlusion happens during the first frames. 
Our HRBC policy performs best compared to other methods.

\noindent {\it Robustness to new object instances.}
We evaluate the robustness of methods to the substitution of a cube by other objects not seen during the training of UR5-Bowl task.
Figure~\ref{fig:bcrl_bowl_unseen} shows the success rate of RLBC and other methods with respect to the number of new objects in simulation. The novel objects are ordered by their dissimilarity with the cube.
The difficulty of grasping unseen objects degrades the performance of grasping skills. In contrast to other methods RLBC is able to automatically recover from errors by making several grasping attempts. 
Table~\ref{tab:hrlbc_bowl_ablation_real} reports corresponding results on a real robot where the cube has been replaced by 10 unseen objects. Similar to other perturbations we observe superior performance of RLBC.

\noindent{\it Impact of the distance between objects.}
We vary the distance between a bottle and a cup in the UR5-Breakfast task. The
smaller distance between objects A and B implies higher probability of collision
between a robot and A when grasping B behind A. The choice of the grasping order
becomes important in such situations. While our method is able to learn the
appropriate grasping order to maximize the chance of completing the task, the BC-ordered and other baselines use pre-defined order.
Figure~\ref{fig:bcrl_breakfast_distance} demonstrates the performance of RLBC and BC-ordered for different object distances in the simulated UR5-Breakfast task. As expected, RLBC learns the correct grasping order and avoids most of collisions. The performance of BC-ordered strongly degrades with the decreasing distance.
In the real-world evaluation, both RLBC and ordered skills succeed in 16 out of 20 episodes when the distance between objects is larger than 10 cm. However, the performance of BC-ordered drops to 8/20 when the cup and the bottle are at $4$cm from each other. In contrast, RLBC chooses the appropriate object to avoid collisions and succeeds in 16 out of 20 trials.

\vspace{-.1cm}

\vspace{-.2cm}
\section{Conclusion}
\vspace{-.2cm}
This paper presents a method to learn combinations of primitive skills.
Our method requires no full-task demonstrations nor intermediate rewards and 
shows excellent results in simulation and on a real robot. We demonstrate the versatility of our approach in challenging settings with dynamic scene changes.
Future work will include learning multiple tasks with shared skills
and addressing contact-rich tasks.

\bibliographystyle{IEEEtran}
\bibliography{IEEEabrv,egbib}

\end{document}